# Structured Noise Modeling for Enhanced Time-Series Forecasting⋆

Sepideh Koohfar[a]

[a]*University of New Hampshire, Durham, 03824, NH, Sepideh.Koohfar@unh.edu*

ABSTRACT

Time-series forecasting remains difficult in real-world settings because temporal patterns operate at multiple scales, from broad contextual trends to fast, fine-grained fluctuations that drive critical decisions. Existing neural models often struggle to represent these interacting dynamics, leading to unstable predictions and reduced reliability in downstream applications. This work introduces a forecast–blur–denoise framework that improves temporal fidelity through structured noise modeling. The approach incorporates a learnable Gaussian Process module that generates smooth, correlated perturbations, encouraging the forecasting backbone to operate as a blur–denoise framework while a dedicated refinement model restores high-resolution temporal detail. Training the components jointly enables natural competence division and avoids the artifacts commonly produced by isotropic corruption methods. Experiments across electricity, traffic, and solar datasets show consistent gains in multi-horizon accuracy and stability. The modular design also allows the blur–denoise layer to operate as a lightweight enhancement for pretrained models, supporting efficient adaptation in limited-data scenarios. By strengthening the reliability and interpretability of fine-scale temporal predictions, this framework contributes to more trustworthy AI systems used in forecasting-driven decision support across energy, infrastructure, and other time-critical domains.

## 1. Introduction

Time-series forecasting is a foundational technology with broad impact across key domains, including economics Capistrán et al. (2010), healthcare Lim (2018), energy and demand planning Salinas et al. (2020), and autonomous systems such as transportation and mobility Chang et al. (2019). Despite substantial progress in neural forecasting models, accurately capturing both long-term dynamics and fine-grained temporal fluctuations remains challenging. Real-world sequences often exhibit multi-scale temporal patterns, where coarse trends and subtle local variations jointly influence future behavior. Recent studies highlight that forecasting models may trade off global structure for local fidelity and vice-versa, motivating architectures tailored for multi-frequency temporal structures Liu et al. (2023); Zeng et al. (2023).

Denoising-based generative models have shown remarkable success in learning structured corruption and recovery processes, particularly in computer vision Ho et al. (2020); Vincent et al. (2010). These approaches typically corrupt data with noise and train a model to recover the underlying clean signal. Motivated by their ability to reconstruct fine-scale details, recent work has begun exploring diffusion-based techniques for time-series modeling and forecasting Rasul et al. (2021); Kollovieh et al. (2023); Shen et al. (2024). In this work, we rethink ideas from denoising models in applications to time series forecasting task, which is defined as follows:

*Time-series forecasting task.* We define a time series as a sequence of observations $\{\chi_t, \gamma_t\}_t$ over time steps $t$. The forecasting task involves predicting values of the target variable $\gamma$ for the next $\tau$ time steps into the future (from $t_0$ to $t_0 + \tau$), given historical $\kappa$ time series observations prior to a cutoff time step $t_0$. We refer to the given historical time series as $X = \{\chi_t, \gamma_t\}_{t=t_0-\kappa}^{t_0}$, and the to be predicted as $Y = \{\gamma_t\}_{t=t_0}^{t_0+\tau}$. Each variable $\chi_t$ consists of $d_\chi$ features including 1) time-based features such as time of the day, season of the year, and 2) other stochastic features observed at time step $t$. Each target variable $\gamma_t$ consists of $d_\gamma \geq 1$ variable(s) of interests, yielding $X_t \in \mathbb{R}^{d_\chi + d_\gamma}$, and $Y_t \in \mathbb{R}^{d_\gamma}$.

Most diffusion-based forecasting methods apply corruption processes primarily adapted from image generation, where noise is often isotropic and uncorrelated across pixels. When applied to time-series, such perturbations can introduce unnatural jitter (see red function in Figure 1 a), making refinement efforts focus on denoising artifacts rather than reconstructing meaningful fine-grained temporal structure.

Motivated by these developments, we introduce a forecast–blur–denoise framework that explicitly partitions responsibilities between coarse forecasting and fine-grained reconstruction. Our method employs a learnable Gaussian Process (GP) blur module to generate temporally correlated distortions of the initial forecast. This structured noise encourages a division of labor: the forecasting backbone concentrates on coarse behavior while the denoiser specializes in restoring high-frequency details. We hypothesize that the GP prior enables smooth, realistic temporal perturbations, avoiding the jittering effect of isotropic noise and leading to more effective competence partitioning (see red function in Figure 1 b).

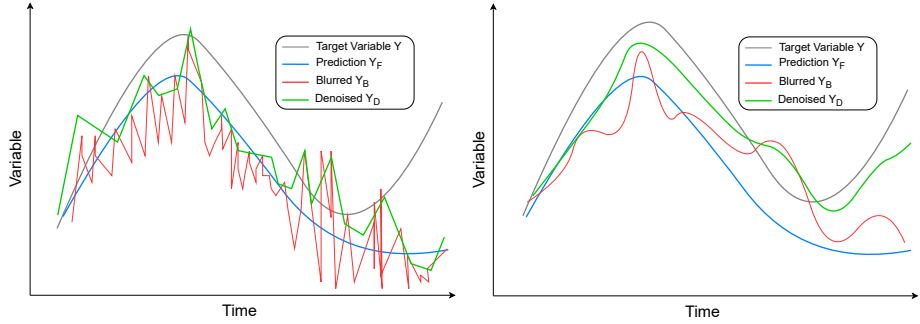

**Figure 1:** A schematic illustration of the hypothesized effect of blurring and denoising on forecasts. In (a), isotropic Gaussian noise leads to high-frequency jitter, prompting the denoiser to suppress artificial fluctuations. In (b), a Gaussian-process blur produces smooth, temporally correlated perturbations, allowing the denoiser to recover fine-grained patterns. This conceptual example highlights the expected qualitative behavior motivating our approach.

Although our experiments evaluate end-to-end training, our framework is modular and can also adapt pretrained forecasting models. In this setting, a pretrained backbone produces an initial coarse forecast, and our blur–denoise refinement module is fine-tuned on the target domain to enhance fine-grained temporal structure. This enables our method to serve as a lightweight adapter layer for zero-shot and few-shot forecasting models, improving granularity without retraining the base model. As time-series foundation models continue to evolve Das et al. (2024); Ansari et al. (2024); Woo et al. (2024), this flexibility allows our approach to complement and enhance modern architectures without modifying their core design.

In this work, we explore a complementary direction for improving time-series forecasting by explicitly separating coarse- and fine-grained prediction capabilities. Our key contributions are as follows:

1. **Forecast–blur–denoise framework.** We propose an end-to-end forecast–blur–denoise architecture that encourages a separation of responsibilities between a coarse-scale forecaster and a fine-scale denoiser. In contrast to strategies that introduce noise only during training, we blur and denoise predictions during both training and inference, hypothesizing that this leads to more consistent refinement behavior.

2. **Model-agnostic refinement layer.** The proposed framework is modular and can be applied on top of a wide range of forecasting models. This design allows our blur–denoise component to operate as a lightweight, architecture-agnostic refinement module.

3. **Alternative to training-only denoising and boosting.** Our approach provides a principled alternative to methods that use denoising solely as a training regularizer or rely on residual boosting. Through controlled experiments, we observe that our integrated forecast–blur–denoise strategy frequently yields improved performance relative to these baselines.

4. **Structured noise hypothesis.** We further hypothesize that for smooth time-series data, structured noise—such as that generated by a Gaussian-process blur model—could be more suitable than isotropic Gaussian perturbations, as it better reflects temporal correlations. Empirical results support this hypothesis across multiple datasets and architectures.

## 2. Related Work

### 2.1. Neural Time-Series Forecasting Models

Deep neural networks have driven significant progress in time-series forecasting, particularly transformer-based architectures Li et al. (2019); Fan et al. (2019). Informer Zhou et al. (2021) introduces ProbSparse attention to reduce computational cost, while Autoformer Wu et al. (2021) incorporates series decomposition and auto-correlation mechanisms to better capture long-range temporal dependencies. These architectures demonstrate strong performance on multi-horizon forecasting tasks and remain common baselines in recent benchmark studies. Our framework is complementary to these models, and we evaluate our refinement method using transformer backbones such as Autoformer and Informer.

Beyond transformers, DeepAR Salinas et al. (2020) generates probabilistic forecasts by modeling parameters of a Gaussian distribution via recurrent neural networks. Despite their effectiveness, RNN-based approaches such as LSTM Hochreiter and Schmidhuber (1997) and GRU have been shown to underperform transformer-based methods in long-range forecasting Koohfar and Dietz (2022). In parallel, linear forecasting approaches such as DLinear Zeng et al. (2023) have recently shown that simple architectures can serve as strong baselines when paired with appropriate trend and seasonal decomposition, motivating the importance of structured inductive biases rather than network depth alone.

## 2.2. Denoising and Diffusion Models for Time Series

Denoising has emerged as a powerful paradigm in generative modeling and forecasting. Early methods such as TimeGrad Rasul et al. (2021) apply denoising diffusion models to time series by reversing Gaussian noise perturbations through recurrent networks. More recent diffusion-based forecasting frameworks demonstrate improved performance and calibration by modeling structured denoising trajectories. Kollovieh et al. introduce self-guided probabilistic diffusion for forecasting Kollovieh et al. (2023), and Shen et al. explore multi-resolution temporal refinement Shen et al. (2024). Surveys Meijer and Chen (2024); Wang et al. (2025) highlight a trend toward time-series-specific adaptations of diffusion processes, including multi-scale representations and spatiotemporal extensions.

However, most diffusion models adapt corruption mechanisms from image generation and rely on isotropic Gaussian noise, which can introduce unnatural jitter when applied to sequential data. In contrast, our approach introduces a learnable Gaussian Process (GP) blur module to generate smooth, temporally correlated perturbations. This enables structured competence partitioning: the forecasting backbone models coarse dynamics, while the denoiser reconstructs fine-grained structure.

## 2.3. Foundation Models and Modular Forecasting

The emergence of time-series foundation models such as TimesFM Das et al. (2024), Chronos Ansari et al. (2024), and Moirai Woo et al. (2024) has introduced new zero-shot and few-shot forecasting paradigms. These models leverage large-scale pretraining to generalize across domains, motivating modular refinement methods that can adapt pretrained backbones efficiently. Our framework aligns with this direction by enabling its blur–denoise module to serve as a lightweight refinement layer without retraining the base model, supporting efficient adaptation in zero-shot and few-shot settings.

# 3. Materials and Methods

Time series forecasting models aim to learn temporal dependencies from observed sequences and to generalize these patterns for future predictions. Although deep architectures, particularly transformer-based models, have demonstrated strong performance on long-range forecasting tasks, they still struggle to faithfully reproduce fine-grained temporal structures. Recent analyses show that modern sequence models tend to prioritize low-frequency components and under-estimate high-frequency temporal variation, resulting in over-smoothing and missed short-term structure (Li et al., 2023). Complex non-linear dynamics, distribution shifts, and the heterogeneity of real-world signals often yield residual errors, even when coarse temporal trends are well captured. Increasing the quantity and diversity of training data may partially alleviate this issue, but domain-specific time series are frequently scarce, expensive, or time-limited. In this work, we explore a complementary direction grounded in the principles of denoising generative models and propose an end-to-end **forecast–blur–denoise** framework that explicitly separates coarse- and fine-resolution forecasting capabilities.

## 3.1. Background: Denoising Models and Isotropic Corruption

Denoising models are trained to reverse a corruption process. For this purpose, true data $X$ is corrupted with noise to obtain a corrupted version $\tilde{X}$. The denoising part of the model is trained to, given $\tilde{X}$, predict the original data $X$. Often this is modelled by predicting the noise, which is then subtracted from $\tilde{X}$ to obtain the original data $X$. Denoising models can be trained for the iterative reversal of noise (reasoning on a series of latents with different noise level) or in single step with a conditional model $X|\tilde{X}$ (Deja et al., 2022). The denoising objective can be optimized by itself or in combination with other objectives such as prompted image generation or time series forecasts (Nichol and Dhariwal, 2021).

Most current work uses a simple corruption process that employs an isotropic Gaussian noise model, where $\tilde{X} \sim \mathcal{N}(\tilde{X}; X, \sigma^2)$ that acts independently on different data points, i.e., pixels for image generation Nichol and Dhariwal (2021) or time steps for time series forecasts (Rasul et al., 2021).

### 3.2. Limitations of Independent Noise for Time Series

Isotropic Gaussian noise is one of the most commonly employed corruption processes, which provides noise that is identically and independently distributed (*i.i.d.*) and when used to generate observations for time series are lacking smooth behavior over time. While isotropic Gaussian noise corruption can (and are) applied to time series, the result is a corruption in the form of jitters (as depicted in red in Figure 1a). However, time series forecasting models are based on the assumption that data points are correlated over time—and hence not i.i.d. The effect on the training problem is that the denoiser may only learn to remove jitters. However, most errors in forecasting models are not due to jitters, as predicted forecasts are usually smooth functions that merely exhibit incorrect fine-grained behavior. In this work we go even one step further and train a smooth and correlated noise/blur model. We hypothesize that this model is more advantageous as it directs the denoising model to focus on accurately predicting the fine-grained details that are locally blurred, rather than simply eliminating jitters.

### 3.3. Temporally Correlated Blur via Gaussian Processes

To obtain an improved fine-grained forecasting model, we instead propose to include a noise/blur model that generates smooth temporally-correlated functions, to train the denoising model, as depicted in Figure 1b. We employ a Gaussian Process (GP) which models the correlation between consecutive samples in a sequence of observations via a kernel function $k_\psi$.

We generate smoothly correlated signals by drawing samples from the probability density function (PDF) of Gaussian Process (GP) denoted as $b_\psi$:

$$b_\psi(\tilde{X} \mid X) = \mathcal{N}\big(X, \, k_\psi(X, X) + \sigma^2 \mathbf{I}\big).$$

GP sample colection produces locally coherent perturbations that primarily affect high-frequency components while preserving coarse temporal trends. Whereas independent Gaussian corruption may encourage the denoiser to focus on suppressing synthetic fluctuations, we hypothesize that temporally structured perturbations better reflect realistic deviations encountered in forecasting settings, and therefore may help the model emphasize reconstruction of meaningful fine-scale structure. As we show in Section 4, empirical results support this hypothesis, indicating that GP-based blur can lead to more effective refinement behavior compared to isotropic Gaussian noise in our evaluation setting.

### 3.4. Forecast–Blur–Denoise Framework

A variety of strategies exist for incorporating denoising within time-series forecasting, including the use of corrupted negative examples, auxiliary denoising objectives, auto-encoder-style representations, and fully integrated end-to-end architectures. Building on preliminary findings, we adopt the integrated approach and develop a unified forecast–blur–denoise framework comprising three components: (1) a forecasting model, (2) a Gaussian-process blur model, and (3) a denoising model (Figure 2). These modules are trained jointly to optimize mean-squared error, and we describe each in turn.

1. **Forecasting model ($\phi$).** Any time series forecasting model can be used here that, given the observations $X = \{x_t; \gamma_t\}_{t=t_0-\tau}^{t_0}$ predicts the future target variables $Y_F$, as represented by the blue box in Figure 2. We experimentally demonstrate that our end-to-end framework will help train enhanced coarse and fine-grained forecasting models. We refer to the set of parameters of the forecasting model as $\phi$.

2. **GP blur model ($\psi$).** The initial predictions $Y_F$ are locally blurred by the mean function of the GP model, depicted as a light red box in Figure 2. As described above, we suggest to use GP as the blur model to obtain $Y_B \sim b_\psi(Y_B|Y_F)$. GP parameters $\psi$ are trained jointly with other parameters of our end-to-end forecasting and denoising model. Alternative noise models could be used here, which we explore in Section 4.

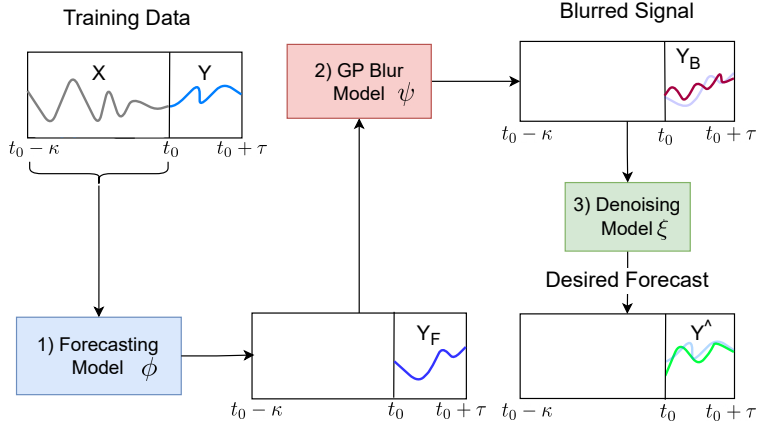

**Figure 2:** Our proposed model's framework for end-to-end training the forecasting, GP blur, and denoising model. The multi-step neural network forecasting model predicts $Y_F$ from historical observations $X$. The predictions $Y_F$ are then blurred by a trained GP model to obtain $Y_B$. The blurred predictions $Y_B$ are then denoised by our denoising model to obtain $Y_D = \hat{Y}$.

3. **Denoising model ($\xi$).** Given the locally blurred predictions $Y_B$, the denoising model focuses on predicting the fine-grained details by removing the blurs, which consequently improves the initial predictions. While many architectures could be chosen for the denoising model, we choose to use the same time series forecasting model with a new set of parameters $\xi$ as the denoiser to obtain final predictions $Y_D = \hat{Y}$. The denoising model is represented by the green box in Figure 2.

The result is a compound model that encourages the initial forecasting model to focus on modelling coarse-grained behavior, and a denoising model that corrects the fine-grained details. This is encouraged by the GP model that will "blur" fine-grained details in the forecast, and a denoising model that focuses on correcting these fine-grained details. Additionally, the denoising component acts as a fall-back for when the initial forecasting model fails, reducing the likelihood of catastrophic errors.

Note that for fixed training data $X$ and $Y$, a new blurring effect is sampled in every epoch, deterring the model from overfitting to any particular blurring pattern.

To optimize the GP model's parameters, we employ a strategy reminiscent of the scalable variational GP method introduced by Hensman et al. (2015) as explained in 3.5. Their scalable variational GP technique offers a computationally efficient approximation of the GP model, achieving nearly linear computational complexity for $k_\psi(X, X)$ as the forecasting horizon increases.

### 3.5. Scalable Variational Gaussian Processes

Naively modeling full GP covariance incurs quadratic complexity in sequence length. To ensure tractability, we employ the variational inducing-point approximation of Hensman et al. (2015), which introduces a learned set of pseudo-inputs $\bar{X}$ and constructs:

$$k_\psi(X, X) \approx k_\psi(X, \bar{X}) \, k_\psi(\bar{X}, \bar{X})^{-1} \, k_\psi(\bar{X}, X)^\top.$$

This approximation enables stochastic mini-batch optimization and yields near-linear scaling with respect to the forecasting horizon, making GP-based blur feasible in modern deep forecasting pipelines.

### 3.6. Training Objective

We train all components jointly via a composite objective:

$$\mathcal{L} = L_{\text{MSE}}(\hat{Y}, Y) \, + \, \lambda \, L_{\text{ELBO}}(\psi).$$

where $L_{\text{MSE}}$ supervises the end-to-end forecasting accuracy and $L_{\text{ELBO}}$ optimizes the GP kernel parameters under the variational evidence lower bound. Following Nichol and Dhariwal (2021), we set $\lambda = 0.001$ to regulate the GP influence such that blur remains a targeted inductive bias rather than dominating learning. This training formulation encourages systematic partitioning of responsibilities: the base model learns smooth coarse structure, while the denoiser specializes in reconstructing high-resolution details.

**Table 1**
Overall results of the quantitative evaluation of our forecast-blur-denoise and other baseline models in terms of average and standard error of **MSE**. We compare the forecasting models on all three datasets with different number of forecasting steps. A lower **MSE** indicates a better model. Our forecast-blur-denoise with GPs enhances the performance of the original Autoformer and isotropic Gaussian noise model (AutoDI). Note that for a fair comparison, all baseline models share the same experimental setup as our proposed model. Reported results may differ from the original baseline papers, and the baseline models are available in our online repository.

| Dataset | Horizon | **AutoDG** | Autoformer | AutoDI | NBeats | DLinear | DeepAR | CMGP | ARIMA |
|---|---|---|---|---|---|---|---|---|---|
| Traffic | 24 | **0.392** ±0.006 | 0.412 ±0.006 | 0.405 ±0.003 | 0.475 ±0.008 | 0.553 ±0.000 | 0.888 ±0.000 | 0.824 ±0.000 | 1.436 ±0.000 |
| | 48 | **0.387** ±0.001 | 0.422 ±0.004 | 0.416 ±0.001 | 0.462 ±0.012 | 0.547 ±0.000 | 0.944 ±0.000 | 0.828 ±0.000 | 1.444 ±0.000 |
| | 72 | **0.380** ±0.001 | **0.383** ±0.003 | 0.394 ±0.002 | 0.465 ±0.003 | 0.540 ±0.000 | 0.877 ±0.000 | 0.893 ±0.000 | 1.459 ±0.000 |
| | 96 | **0.385** ±0.003 | 0.400 ±0.004 | 0.411 ±0.002 | 0.464 ±0.002 | 0.539 ±0.000 | 0.860 ±0.000 | 0.859 ±0.000 | 1.444 ±0.000 |
| Electricity | 24 | **0.165** ±0.001 | 0.187 ±0.003 | 0.170 ±0.001 | 0.200 ±0.001 | 0.222 ±0.000 | 1.039 ±0.000 | 1.000 ±0.000 | 1.707 ±0.000 |
| | 48 | **0.188** ±0.003 | 0.203 ±0.008 | 0.207 ±0.003 | 0.218 ±0.000 | 0.238 ±0.000 | 1.014 ±0.000 | 0.987 ±0.000 | 1.729 ±0.000 |
| | 72 | **0.209** ±0.004 | 0.230 ±0.001 | 0.253 ±0.004 | 0.234 ±0.007 | 0.264 ±0.000 | 1.023 ±0.000 | 0.993 ±0.000 | 1.759 ±0.000 |
| | 96 | **0.211** ±0.001 | 0.230 ±0.014 | 0.316 ±0.002 | 0.237 ±0.001 | 0.264 ±0.000 | 1.013 ±0.000 | 0.971 ±0.000 | 1.747 ±0.000 |
| Solar | 24 | **0.446** ±0.002 | 0.472 ±0.003 | 0.473 ±0.001 | 0.612 ±0.006 | 0.828 ±0.000 | 0.999 ±0.000 | 1.001 ±0.000 | 1.869 ±0.000 |
| | 48 | **0.546** ±0.003 | 0.603 ±0.004 | 0.574 ±0.001 | 0.717 ±0.001 | 0.928 ±0.000 | 0.968 ±0.000 | 1.007 ±0.000 | 1.872 ±0.000 |
| | 72 | **0.666** ±0.003 | **0.667** ±0.004 | 0.698 ±0.002 | 0.766 ±0.006 | 0.978 ±0.000 | 0.974 ±0.000 | 1.002 ±0.000 | 1.855 ±0.000 |
| | 96 | **0.713** ±0.004 | 0.739 ±0.009 | 0.730 ±0.005 | 0.827 ±0.001 | 1.004 ±0.000 | 0.974 ±0.000 | 0.997 ±0.000 | 1.874 ±0.000 |

We now detail the experimental protocol used to evaluate our proposed approach.

### 3.7. Datasets

We conduct empirical studies on three canonical benchmark datasets that have been extensively adopted in contemporary time–series forecasting research (Salinas et al., 2020; Li et al., 2019; Wu et al., 2021; Zhou et al., 2021). These datasets represent diverse real-world settings and are widely recognized for benchmarking multi-horizon forecasting systems.

**Traffic:** Hourly traffic occupancy rates ($y_t \in [0, 1]$) collected from 440 loop sensors deployed on freeways in the San Francisco Bay Area. This dataset characterizes complex spatiotemporal congestion dynamics.

**Solar Energy:** Hourly measurements of photovoltaic power generation across multiple geographically dispersed sites in the United States, capturing heterogeneous load profiles and diurnal/seasonal demand fluctuations.

**Electricity:** Hourly electricity consumption records from 370 residential and commercial clients, reflecting highly nonstationary renewable-energy dynamics influenced by meteorological variability.

**Table 2**
Comparison of different denoising baselines to our forecast-blur-denoise approach with GPs when treating Autoformer forecasting model. We find that our approach predominantly outperforms the other denoising approaches. Results are reported as average and standard error of **MSE**. A lower **MSE** indicates a better forecasting model.

| Dataset | Horizon | **AutoDG** | Autoformer | AutoDI | AutoDWC | AutoRB | AutoDT |
|---|---|---|---|---|---|---|---|
| Traffic | 24 | **0.392** ±0.006 | 0.412 ±0.006 | 0.405 ±0.003 | 0.400 ±0.005 | 0.447 ±0.006 | 0.430 ±0.015 |
| | 48 | **0.387** ±0.001 | 0.422 ±0.007 | 0.416 ±0.007 | 0.417 ±0.009 | 0.450 ±0.005 | 0.410 ±0.005 |
| | 72 | **0.380** ±0.001 | **0.383** ±0.002 | 0.394 ±0.002 | 0.398 ±0.003 | 0.430 ±0.004 | 0.404 ±0.006 |
| | 96 | **0.385** ±0.003 | 0.400 ±0.004 | 0.411 ±0.002 | 0.405 ±0.001 | 0.413 ±0.002 | 0.422 ±0.002 |
| Electricity | 24 | **0.165** ±0.001 | 0.187 ±0.003 | 0.170 ±0.001 | 0.174 ±0.00 | 0.260 ±0.001 | 0.170 ±0.007 |
| | 48 | **0.188** ±0.003 | 0.203 ±0.008 | 0.207 ±0.003 | 0.219 ±0.002 | 0.222 ±0.002 | 0.200 ±0.004 |
| | 72 | **0.209** ±0.004 | 0.230 ±0.001 | 0.253 ±0.004 | 0.218 ±0.010 | 0.234 ±0.022 | 0.212 ±0.002 |
| | 96 | **0.211** ±0.001 | 0.230 ±0.014 | 0.316 ±0.002 | 0.226 ±0.008 | 0.296 ±0.011 | 0.218 ±0.004 |
| Solar | 24 | **0.446** ±0.002 | 0.472 ±0.003 | 0.473 ±0.006 | 0.449 ±0.003 | 0.527 ±0.006 | 0.457 ±0.004 |
| | 48 | **0.546** ±0.005 | 0.603 ±0.003 | 0.574 ±0.001 | 0.605 ±0.005 | 0.595 ±0.005 | 0.598 ±0.003 |
| | 72 | **0.666** ±0.003 | **0.667** ±0.004 | 0.698 ±0.002 | 0.690 ±0.010 | 0.718 ±0.002 | 0.670 ±0.006 |
| | 96 | **0.713** ±0.004 | 0.739 ±0.009 | 0.730 ±0.005 | 0.732 ±0.006 | 0.753 ±0.007 | 0.733 ±0.006 |

For each dataset, we sample approximately 40,000 input-target windows, each consisting of $\kappa = 192$ historical time steps and a forecasting horizon of $\tau \in \{24, 48, 72, 96\}$ future steps. All series are standardized via Z-score normalization. We adopt an 80%/10%/10% train/validation/test split across samples.

### 3.8. Evaluation metrics.

We evaluate forecasting quality using mean squared error (MSE), and mean absolute error (MAE):

$$\text{MSE} = \frac{1}{n} \sum_{t=1}^{n} (\mathbf{y}_t - \hat{\mathbf{y}}_t)^2, \text{MAE} = \frac{1}{n} \sum_{t=1}^{n} |\mathbf{y}_t - \hat{\mathbf{y}}_t|.$$

where $n$ denotes the forecast horizon. MAE-based results exhibit identical trends and are reported in Appendix 7.

### 3.9. Models and Baselines

our forecast–blur–denoise framework is model-agnostic, we evaluate it in conjunction with two established transformer-based forecasting baselines widely adopted in the literature:

**Autoformer (Wu et al., 2021):** Exploits autocorrelation-based decomposition and seasonal-trend separation mechanisms.

**Informer (Zhou et al., 2021):** Utilizes ProbSparse attention to scale long-sequence modeling efficiently.

We evaluate the following configurations:

**AutoDG/InfoDG (ours):** Forecast–blur–denoise with a learnable GP blur module.

**AutoDI/InfoDI:** Isotropic-Gaussian blur instead of GP blur (controlled denoising baseline).

**Autoformer/Informer:** Original forecasting backbone (no denoising).

We further benchmark against a representative suite of widely-adopted forecasting baselines:

**1) ARIMA (Hyndman and Khandakar, 2008):** Autoregressive integrated moving-average model.

**2) CMGP (Chakrabarty et al., 2021):** Calibration model leveraging Bayesian optimization with GP priors.

**3) DeepAR (Salinas et al., 2020):** Probabilistic RNN-based forecaster.

**4) DLinear (Zeng et al., 2023):** Linear MLP-based architecture.

**5) NBEATS (Oreshkin et al., 2020):** Deep residual framework decomposing trend, seasonality, and residual components.

To rigorously evaluate design choices, we perform systematic ablation studies analyzing the influence of GP blur, isotropic noise corruption, and denoising strategies:

**AutoDWB/InfoDWB:** Denoising without blur (forecast–denoise only).

**AutoRB/InfoRB:** Residual boosting: a second model trained on forecast residuals.

**AutoDT/InfoDT:** GP-blur-guided denoising applied only during training (no blur at inference).

### 3.10. Training Procedure and Hyperparameters

All models are trained using multiple random seeds. Hyperparameters are optimized via Optuna (Akiba et al., 2019). We search over latent dimensionalities $\{16, 32\}$, layer counts $\{1, 2\}$, and learning-rate warm-up steps $\{1000, 8000\}$ following Vaswani et al. (2017). For isotropic denoising, the noise scale $\sigma$ is learned within $[0, 0.1]$.

Transformer-based models employ eight-head multi-head attention. Gaussian processes are implemented via `ApproximateGP` from GPyTorch. We use the Adam optimizer (Kingma and Ba, 2015). Training is conducted with batch size 256 for 50 epochs. Experiments are executed on a single NVIDIA A40 GPU (45GB RAM), with each training epoch requiring approximately 25 seconds.

## 4. Results

Tables 1 and 3 report the quantitative performance of our framework applied to the Autoformer and Informer architectures, alongside widely-adopted forecasting baselines on the Electricity, Traffic, and Solar benchmarks. Results are expressed as mean squared error (MSE) with standard errors across multiple random seeds and forecasting horizons $\tau \in \{24, 48, 72, 96\}$. For ease of comparison, baseline results are included in both tables.

Across all datasets and horizons, the GP-based forecast–blur–denoise variants (AutoDG/InfoDG) consistently achieve lower error than both the original forecasting models and the isotropic denoising variant (AutoDI/InfoDI). Performance gains are especially pronounced under longer horizons, where modeling high-frequency structure becomes increasingly challenging. Figure 3 corroborate these results, illustrating smoother predictive trajectories and reduced deviation from ground truth.

We additionally benchmark against a representative suite of established forecasting baselines, including ARIMA, CMGP, DeepAR, DLinear, and NBEATS. While these models exhibit competitive performance under specific settings, they consistently underperform relative to the GP-enhanced architectures, affirming the benefit of structured blur-driven refinement.

Finally, we evaluate ablation variants in Tables 2 and 4. The denoiser-only configuration (AutoDWB/InfoDWB) provides limited gains and frequently fails to reconstruct high-frequency structure. Similarly, residual boosting (AutoRB/InfoRB) does not consistently translate into improved predictive fidelity. The training-only blur configuration (AutoDT/InfoDT) also performs worse than the full strategy, demonstrating that applying blur during inference plays a critical role in enabling effective coarse-to-fine forecast refinement.

## 5. Discussion

Figure presents qualitative forecasts for a 72-step horizon, illustrating the behavior of different variants. The untreated Autoformer tracks global trends but struggles with subtle temporal structures. The isotropic denoising approach introduces high-frequency jitter, while denoising without blur fails to recover smooth transitions and extreme values. In contrast, the GP-based blur mechanism yields predictions that accurately capture both coarse trajectories and fine-scale variations, reflecting improved fidelity in slope changes, peak behavior, and local dynamics.

These results highlight the central value of temporally-correlated blur as an inductive bias. By introducing structured, smooth perturbations aligned with the stochastic properties of real-world time series, the GP prior facilitates a principled division of responsibilities: the forecasting backbone models long-range dependencies and global patterns, while the denoising network specializes in reconstructing fine-grained temporal features. This competence separation consistently yields enhanced accuracy and smoother multi-horizon forecasts.

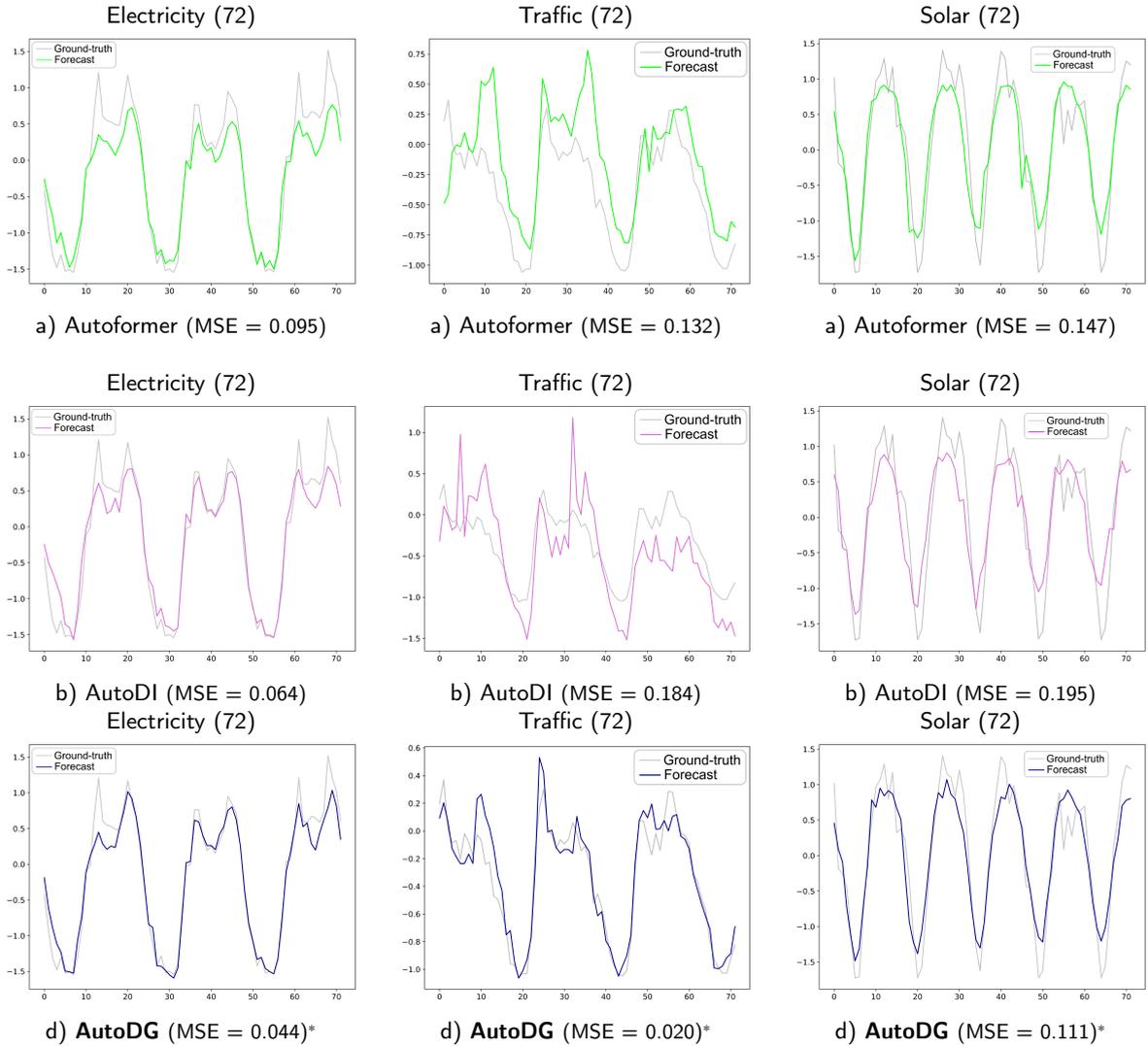

**Figure 3:** Example forecasts of four treatments of autoformer model on Traffic, Electricity, and Solar datastes for 72 future time steps. The values are plotted in Z-score normalized space. Best MSE results for each column (dataset) are denoted with the ∗ symbol.

The comparative performance of AutoDI/InfoDI underscores the importance of noise structure. Independent Gaussian corruption, while effective in high-dimensional image domains, is misaligned with temporal continuity and can introduce artifacts rather than foster meaningful correction. Conversely, GP-induced blur reflects realistic uncertainty and enables effective refinement.

Additional MAE results and extended ablation analyses are included in Appendix 7, along with qualitative visualizations and per-horizon error curves. These supplementary results further reinforce the robustness and generality of the proposed approach across architectures, datasets, and prediction lengths.

**Table 3**
Overall results of the quantitative evaluation of our forecast-blur-denoise and other baseline models in terms of average and standard error of **MSE**. We compare the forecasting models on all three datasets with different number of forecasting steps. A lower **MSE** indicates a better model. Our forecast-blur-denoise with GPs enhances the performance of the original Informer and isotropic Gaussian noise model (InfoDI). Note that for a fair comparison, all baseline models share the same experimental setup as our proposed model. Reported results may differ from the original baseline papers, and the baseline models are available in our online repository.

| Dataset | Horizon | **InfoDG** | Informer | InfoDI | NBeats | DLinear | DeepAR | CMGP | ARIMA |
|---|---|---|---|---|---|---|---|---|---|
| Traffic | 24 | **0.398** | 0.421 | 0.415 | 0.475 | 0.553 | 0.888 | 0.824 | 1.436 |
| | | ±0.006 | ±0.006 | ±0.003 | ±0.008 | ±0.000 | ±0.000 | ±0.000 | ±0.000 |
| | 48 | **0.399** | 0.434 | 0.395 | 0.462 | 0.547 | 0.944 | 0.828 | 1.444 |
| | | ±0.001 | ±0.004 | ±0.001 | ±0.012 | ±0.000 | ±0.000 | ±0.000 | ±0.000 |
| | 72 | **0.380** | 0.436 | 0.395 | 0.465 | 0.540 | 0.877 | 0.893 | 1.459 |
| | | ±0.001 | ±0.001 | ±0.002 | ±0.002 | ±0.000 | ±0.000 | ±0.000 | ±0.000 |
| | 96 | **0.397** | **0.402** | **0.402** | 0.464 | 0.539 | 0.860 | 0.859 | 1.444 |
| | | ±0.003 | ±0.003 | ±0.004 | ±0.004 | ±0.000 | ±0.000 | ±0.000 | ±0.000 |
| Electricity | 24 | **0.193** | 0.222 | 0.212 | 0.200 | 0.222 | 1.039 | 1.000 | 1.707 |
| | | ±0.001 | ±0.001 | ±0.003 | ±0.001 | ±0.000 | ±0.000 | ±0.000 | ±0.000 |
| | 48 | **0.222** | 0.262 | 0.229 | 0.218 | 0.238 | 1.014 | 0.987 | 1.729 |
| | | ±0.003 | ±0.007 | ±0.003 | ±0.003 | ±0.000 | ±0.000 | ±0.000 | ±0.000 |
| | 72 | **0.238** | 0.280 | 0.253 | **0.234** | 0.264 | 1.023 | 0.993 | 1.759 |
| | | ±0.001 | ±0.004 | ±0.004 | ±0.007 | ±0.000 | ±0.000 | ±0.000 | ±0.000 |
| | 96 | 0.242 | 0.289 | 0.275 | **0.237** | 0.264 | 1.013 | 1.130 | 1.747 |
| | | ±0.001 | ±0.002 | ±0.014 | ±0.001 | ±0.000 | ±0.000 | ±0.000 | ±0.000 |
| Solar | 24 | **0.455** | 0.524 | **0.465** | 0.612 | 0.828 | 0.999 | 0.971 | 1.869 |
| | | ±0.009 | ±0.003 | ±0.006 | ±0.006 | ±0.000 | ±0.000 | ±0.000 | ±0.000 |
| | 48 | **0.556** | 0.629 | 0.570 | 0.717 | 0.928 | 0.968 | 1.007 | 1.872 |
| | | ±0.005 | ±0.003 | ±0.005 | ±0.001 | ±0.000 | ±0.000 | ±0.000 | ±0.000 |
| | 72 | **0.643** | 0.729 | 0.707 | 0.766 | 0.978 | 0.974 | 1.002 | 1.855 |
| | | ±0.003 | ±0.023 | ±0.002 | ±0.006 | ±0.000 | ±0.000 | ±0.000 | ±0.000 |
| | 96 | **0.708** | 0.770 | 0.766 | 0.827 | 1.004 | 0.974 | 0.997 | 1.874 |
| | | ±0.004 | ±0.004 | ±0.009 | ±0.005 | ±0.000 | ±0.000 | ±0.000 | ±0.000 |

## 6. Conclusion

In this work, we addressed the multi-horizon time-series forecasting problem and introduced an end-to-end forecast–blur–denoise framework. By jointly training a forecasting backbone, a Gaussian-process (GP) blur module, and a denoising network, our approach enforces a principled separation of responsibilities: the base forecaster learns coarse-grained temporal dynamics, while the denoiser specializes in reconstructing fine-grained structure. Leveraging variational GP optimization enables the blur process to adaptively learn smooth, temporally-correlated perturbations that align with real-world stochastic behavior, ultimately improving fidelity at multiple temporal resolutions.

Contrary to conventional denoising approaches that rely on isotropic Gaussian corruption, we hypothesized that structured, correlated noise is better suited to sequential data. Consistent with prior insights on temporally-aware noise models (Robinson et al., 2018), our empirical results confirm that GP-based blur yields substantial benefits over both uncorrelated noise and no-blur settings. Across three benchmark datasets and multiple forecasting horizons, the proposed method delivers consistently lower error than the underlying forecasting models, isotropic denoising baselines, and training-only noise variants.

A key advantage of our approach is its model-agnostic design. While our experiments instantiate the framework with commonly-benchmarked transformer forecasters such as Autoformer and Informer, the method is agnostic to backbone architecture and can be readily integrated with emerging pretrained time-series models. This provides a promising direction for future research, where our blur-denoise refinement module may serve as an adapter layer to enhance fine-scale accuracy in zero-shot and few-shot forecasting settings without retraining large base models.

**Table 4**
Comparison of different denoising baselines to our forecast-blur-denoise approach with GPs when treating Informer forecasting model. We find that our approach predominantly outperforms the other denoising approaches. Results are reported as average and standard error of **MSE**. A lower **MSE** indicates a better forecasting model.

| Dataset | Horizon | **InfoDG** | Informer | InfoDI | InfoDWC | InfoRB | InfoDT |
|---------|---------|-----------|----------|--------|---------|--------|--------|
| Traffic | 24 | **0.398**±0.005 | 0.421±0.005 | 0.415±0.002 | 0.406±0.002 | 0.435±0.005 | 0.473±0.003 |
| | 48 | **0.399**±0.004 | 0.434±0.014 | 0.395±0.007 | 0.392±0.003 | **0.395** ±0.011 | 0.421±0.014 |
| | 72 | **0.380**±0.001 | 0.436±0.015 | 0.395±0.001 | 0.392±0.001 | 0.407±0.009 | 0.421±0.011 |
| | 96 | **0.397**±0.003 | 0.402±0.002 | **0.402**±0.006 | **0.394**±0.003 | 0.412±0.007 | 0.414±0.015 |
| Electricity | 24 | **0.193**±0.003 | 0.222±0.006 | 0.212±0.001 | 0.204±0.005 | 0.225±0.009 | 0.230±0.007 |
| | 48 | **0.222**±0.003 | 0.262±0.013 | 0.229±0.003 | 0.241±0.007 | 0.261±0.014 | 0.256±0.004 |
| | 72 | **0.238**±0.001 | 0.280±0.006 | 0.253±0.006 | 0.263±0.013 | 0.262±0.008 | 0.268±0.008 |
| | 96 | **0.242**±0.004 | 0.289±0.011 | 0.275±0.005 | 0.279±0.006 | 0.283±0.001 | 0.275±0.007 |
| Solar | 24 | **0.455**±0.007 | 0.524±0.002 | 0.465±0.006 | 0.457±0.006 | 0.498±0.010 | 0.512±0.012 |
| | 48 | **0.556**±0.011 | 0.629±0.021 | 0.570±0.007 | 0.590±0.016 | 0.623±0.016 | 0.629±0.023 |
| | 72 | **0.643**±0.022 | 0.729±0.024 | 0.707±0.026 | 0.708±0.014 | 0.748±0.010 | 0.726±0.006 |
| | 96 | **0.708**±0.010 | 0.770±0.017 | 0.766±0.006 | 0.739±0.010 | 0.781±0.017 | 0.777±0.000 |

Overall, our findings demonstrate that temporally-correlated blur offers a powerful inductive bias for hierarchical temporal modeling, leading to meaningful gains in coarse and fine-grained forecasting performance. We expect this perspective—explicitly structuring forecast refinement via learned smooth noise priors—to complement and enhance next-generation pretrained forecasting architectures and foundation models for time-series analysis.

# 7. Appendix

Here, we provide the quantitative results in terms of MAE evaluation metric. Table 5 and 7 include the MAE results of the initial three treatments and other baseline models. Note that results of the baseline models are repeated for the purpose of readability. Additionally, the MAE results of the ablation studies of other denoising/boosting models are reported in Table 6 and 8.

To show that the improvement of our proposed model indeed stems from its ability and not the higher number of parameters regarding the GP model, we conducted an ablation study by increasing the number of layers (parameters) of the standalone forecasting models and included the results in terms of both MSE and MAE in Table 9 and 10.

**Table 5**
Overall results of the quantitative evaluation of our forecast-blur-denoise and other baseline models in terms of average and standard error of **MAE**. We compare the forecasting models on all three datasets with different number of forecasting steps. A lower **MAE** indicates a better model. Our forecast-blur-denoise with GPs enhances the performance of the original Autoformer and isotropic Gaussian noise model (AutoDI). Note that for a fair comparison, all baseline models share the same experimental setup as our proposed model. Reported results may differ from the original baseline papers, and the baseline models are available in our online repository.

| Dataset | Horizon | **AutoDG(Ours)** | Autoformer | AutoDI | NBeats | DLinear | DeepAR | CMGP | ARIMA |
|---|---|---|---|---|---|---|---|---|---|
| Traffic | 24 | **0.333** | **0.334** | **0.340** | 0.384 | 0.447 | 0.652 | 0.645 | 0.770 |
| | | ±0.010 | ±0.007 | ±0.007 | ±0.001 | ±0.000 | ±0.000 | ±0.000 | ±0.000 |
| | 48 | **0.328** | 0.368 | 0.343 | 0.408 | 0.462 | 0.650 | 0.642 | 0.776 |
| | | ±0.001 | ±0.006 | ±0.006 | ±0.003 | ±0.000 | ±0.000 | ±0.000 | ±0.000 |
| | 72 | **0.358** | **0.356** | **0.356** | 0.413 | 0.466 | 0.636 | 0.648 | 0.782 |
| | | ±0.013 | ±0.003 | ±0.005 | ±0.004 | ±0.000 | ±0.000 | ±0.000 | ±0.000 |
| | 96 | **0.333** | 0.359 | 0.366 | 0.414 | 0.471 | 0.632 | 0.647 | 0.773 |
| | | ±0.000 | ±0.004 | ±0.004 | ±0.002 | ±0.000 | ±0.000 | ±0.000 | ±0.000 |
| Electricity | 24 | **0.249** | 0.265 | 0.258 | 0.294 | 0.299 | 0.862 | 0.840 | 0.959 |
| | | ±0.001 | ±0.003 | ±0.001 | ±0.004 | ±0.000 | ±0.000 | ±0.000 | ±0.000 |
| | 48 | **0.275** | 0.292 | 0.301 | 0.310 | 0.308 | 0.853 | 0.839 | 0.971 |
| | | ±0.003 | ±0.007 | ±0.003 | ±0.007 | ±0.000 | ±0.000 | ±0.000 | ±0.000 |
| | 72 | **0.303** | **0.297** | **0.303** | 0.322 | 0.323 | 0.856 | 0.836 | 0.987 |
| | | ±0.004 | ±0.006 | ±0.004 | ±0.007 | ±0.000 | ±0.000 | ±0.000 | ±0.000 |
| | 96 | **0.304** | 0.372 | 0.325 | 0.324 | 0.329 | 0.850 | 0.832 | 0.983 |
| | | ±0.001 | ±0.010 | ±0.002 | ±0.002 | ±0.000 | ±0.000 | ±0.000 | ±0.000 |
| Solar | 24 | **0.548** | 0.603 | 0.574 | 0.632 | 0.801 | 0.885 | 0.885 | 1.100 |
| | | ±0.009 | ±0.002 | ±0.008 | ±0.001 | ±0.000 | ±0.000 | ±0.000 | ±0.000 |
| | 48 | **0.612** | 0.656 | 0.638 | 0.710 | 0.864 | 0.865 | 0.888 | 1.102 |
| | | ±0.003 | ±0.003 | ±0.003 | ±0.001 | ±0.000 | ±0.000 | ±0.000 | ±0.000 |
| | 72 | **0.702** | 0.729 | **0.702** | 0.744 | 0.894 | 0.873 | 0.885 | 1.106 |
| | | ±0.0001 | ±0.017 | ±0.001 | ±0.001 | ±0.000 | ±0.000 | ±0.000 | ±0.000 |
| | 96 | **0.725** | 0.754 | 0.747 | 0.781 | 0.911 | 0.866 | 0.882 | 1.097 |
| | | ±0.000 | ±0.009 | ±0.005 | ±0.002 | ±0.000 | ±0.000 | ±0.000 | ±0.000 |

**Table 6**
Comparison of different denoising baselines to our proposed model when treating the Autoformer model. We find that our approach predominantly outperforms the other denoising approaches. Results are reported as average and standard error of **MAE**. A lower **MAE** indicates a better forecasting model.

| Dataset | Horizon | **AutoDG(Ours)** | Autoformer | AutoDI | AutoDWC | AutoRB | AutoDT |
|---|---|---|---|---|---|---|---|
| Traffic | 24 | **0.333**±0.010 | **0.334**±0.007 | **0.340**±0.007 | 0.345±0.009 | 0.391±0.005 | 0.349±0.005 |
| | 48 | **0.328**±0.001 | 0.368±0.006 | 0.343±0.006 | 0.351±0.005 | 0.359±0.002 | 0.361±0.016 |
| | 72 | **0.358**±0.013 | **0.356**±0.003 | **0.356**±0.005 | 0.361±0.002 | 0.383±0.005 | 0.379±0.001 |
| | 96 | **0.304**±0.000 | 0.325±0.004 | 0.372±0.004 | 0.362±0.004 | 0.368±0.005 | 0.379±0.005 |
| Electricity | 24 | **0.249**±0.001 | 0.265±0.003 | 0.258±0.001 | 0.272±0.001 | 0.380±0.001 | 0.263±0.007 |
| | 48 | **0.275**±0.003 | 0.292±0.007 | 0.301±0.003 | 0.306±0.002 | 0.311±0.002 | 0.288±0.003 |
| | 72 | **0.303**±0.004 | **0.297**±0.006 | **0.303**±0.004 | **0.295**±0.009 | 0.330±0.021 | **0.305**±0.003 |
| | 96 | **0.304**±0.001 | 0.372±0.010 | 0.325±0.002 | 0.324±0.005 | 0.386±0.010 | 0.318±0.005 |
| Solar | 24 | **0.548**±0.009 | 0.603±0.002 | 0.574±0.008 | 0.549±0.008 | 0.601±0.005 | 0.598±0.006 |
| | 48 | **0.612**±0.003 | 0.656±0.003 | 0.638±0.003 | 0.656±0.002 | 0.645±0.003 | 0.655±0.003 |
| | 72 | **0.702**±0.001 | 0.729±0.017 | **0.702**±0.001 | 0.724±0.008 | 0.720±0.010 | 0.709±0.004 |
| | 96 | **0.725**±0.000 | 0.754±0.009 | 0.747±0.005 | 0.746±0.006 | 0.738±0.007 | 0.745±0.006 |

**Table 7**
Overall results of the quantitative evaluation of our forecast-blur-denoise and other baseline models in terms of average and standard error of **MAE**. We compare the forecasting models on all three datasets with different number of forecasting steps. A lower **MAE** indicates a better model. Our forecast-blur-denoise with GPs enhances the performance of the original Informer and isotropic Gaussian noise model (InfoDI). Note that for a fair comparison, all baseline models share the same experimental setup as our proposed model. Reported results may differ from the original baseline papers, and the baseline models are available in our online repository.

| Dataset | Horizon | InfoDG(Ours) | Informer | InfoDI | NBeats | DLinear | DeepAR | CMGP | ARIMA |
|---|---|---|---|---|---|---|---|---|---|
| Traffic | 24 | 0.355 ±0.007 | **0.329** ±0.006 | 0.342 ±0.004 | 0.384 ±0.001 | 0.447 ±0.000 | 0.652 ±0.000 | 0.645 ±0.000 | 0.770 ±0.000 |
| | 48 | **0.350** ±0.001 | **0.354** ±0.012 | **0.362** ±0.005 | 0.408 ±0.003 | 0.462 ±0.000 | 0.650 ±0.000 | 0.642 ±0.000 | 0.776 ±0.000 |
| | 72 | **0.345** ±0.003 | 0.377 ±0.002 | 0.353 ±0.008 | 0.413 ±0.004 | 0.466 ±0.000 | 0.636 ±0.000 | 0.648 ±0.000 | 0.782 ±0.000 |
| | 96 | **0.397** ±0.004 | 0.402 ±0.011 | 0.402 ±0.005 | **0.414** ±0.002 | 0.471 ±0.000 | 0.632 ±0.000 | 0.647 ±0.000 | 0.773 ±0.000 |
| Electricity | 24 | **0.290** ±0.008 | **0.300** ±0.005 | **0.298** ±0.003 | **0.294** ±0.004 | **0.299** ±0.000 | 0.862 ±0.000 | 0.840 ±0.000 | 0.959 ±0.000 |
| | 48 | **0.311** ±0.002 | 0.349 ±0.009 | 0.325 ±0.002 | 0.310 ±0.007 | 0.308 ±0.000 | 0.853 ±0.000 | 0.839 ±0.000 | 0.971 ±0.000 |
| | 72 | **0.345** ±0.003 | 0.377 ±0.002 | 0.353 ±0.008 | 0.322 ±0.002 | 0.323 ±0.000 | 0.856 ±0.000 | 0.836 ±0.000 | 0.987 ±0.000 |
| | 96 | **0.342** ±0.004 | 0.378 ±0.011 | 0.379 ±0.005 | 0.324 ±0.002 | 0.329 ±0.000 | 0.850 ±0.000 | 0.832 ±0.000 | 0.983 ±0.000 |
| Solar | 24 | **0.533** ±0.005 | 0.597 ±0.002 | 0.563 ±0.003 | 0.632 ±0.001 | 0.801 ±0.000 | 0.885 ±0.000 | 0.885 ±0.000 | 1.100 ±0.000 |
| | 48 | **0.624** ±0.007 | 0.681 ±0.013 | 0.635 ±0.005 | 0.710 ±0.001 | 0.864 ±0.000 | 0.865 ±0.000 | 0.888 ±0.000 | 1.102 ±0.000 |
| | 72 | **0.690** ±0.013 | 0.752 ±0.017 | 0.735 ±0.019 | 0.744 ±0.001 | 0.894 ±0.000 | 0.873 ±0.000 | 0.885 ±0.000 | 1.106 ±0.000 |
| | 96 | **0.727** ±0.006 | 0.772 ±0.012 | 0.764 ±0.004 | 0.781 ±0.002 | 0.911 ±0.000 | 0.866 ±0.000 | 0.882 ±0.000 | 1.097 ±0.000 |

**Table 8**
Comparison of different denoising baselines to our forecast-blur-denoise approach with GPs when treating Informer forecasting model. We find that our approach predominantly outperforms the other denoising approaches. Results are reported as average and standard error of **MAE**. A lower **MAE** indicates a better forecasting model.

| Dataset | Horizon | InfoDG(Ours) | Informer | InfoDI | InfoDWC | InfoRB | InfoDT |
|---|---|---|---|---|---|---|---|
| Traffic | 24 | 0.355±0.007 | **0.329**±0.006 | 0.342±0.004 | 0.337±0.003 | 0.331±0.003 | 0.379±0.003 |
| | 48 | **0.345**±0.001 | 0.377±0.013 | 0.353±0.005 | 0.348±0.003 | 0.353±0.009 | 0.375±0.006 |
| | 72 | **0.345**±0.003 | 0.377±0.010 | 0.353±0.004 | 0.348±0.001 | 0.379±0.005 | 0.361±0.006 |
| | 96 | **0.350**±0.004 | **0.354**±0.006 | **0.362**±0.007 | **0.348**±0.008 | 0.379±0.000 | **0.361**±0.011 |
| Electricity | 24 | **0.290**±0.008 | **0.300**±0.005 | **0.298**±0.003 | **0.295**±0.004 | **0.302**±0.009 | 0.318±0.003 |
| | 48 | **0.311**±0.002 | 0.349±0.009 | 0.325±0.002 | 0.333±0.004 | 0.343±0.009 | 0.343±0.005 |
| | 72 | **0.336**±0.003 | 0.371±0.002 | 0.359±0.008 | 0.362±0.008 | 0.359±0.006 | 0.367±0.008 |
| | 96 | **0.342**±0.004 | 0.378±0.0011 | 0.379±0.005 | 0.384±0.006 | 0.375±0.001 | 0.370±0.007 |
| Solar | 24 | **0.533**±0.005 | 0.597±0.002 | 0.563±0.003 | 0.551±0.001 | 0.573±0.008 | 0.596±0.007 |
| | 48 | **0.624**±0.007 | 0.681±0.013 | 0.635±0.005 | 0.649±0.011 | 0.675±0.012 | 0.681±0.012 |
| | 72 | **0.690**±0.013 | 0.752±0.017 | 0.735±0.019 | 0.736±0.011 | 0.763±0.020 | 0.735±0.004 |
| | 96 | **0.727**±0.006 | 0.772±0.012 | 0.764±0.004 | 0.753±0.005 | 0.777±0.014 | 0.766±0.002 |

**Table 9**
Comparison of our forecast-blur-denoise approach with standalone forecasting models with higher number of layers (parameters) denoted by † sign. Initially, the number of layers for our proposed model and other baselines are chosen from $\{1, 2\}$, however to show that the performance of our model indeed stems from its mechanism, we included the results of standalone forecasting models with number of layers chosen from $\{3, 4\}$. Results are reported as average and standard error of **MSE**. A lower **MSE** indicates a better forecasting model.

| Dataset | Horizon | AutoDG(Ours) | Autoformer | Autoformer† | InfoDG(Ours) | Informer | Informer† |
|---|---|---|---|---|---|---|---|
| Traffic | 24 | **0.392** ±0.006 | 0.412 ±0.006 | **0.359** ±0.007 | **0.398** ±0.006 | 0.421 ±0.006 | 0.422 ±0.009 |
| | 48 | **0.387** ±0.001 | 0.422 ±0.007 | **0.383** ±0.001 | **0.399** ±0.001 | 0.434 ±0.001 | 0.486 ±0.010 |
| | 72 | **0.380** ±0.001 | **0.383** ±0.002 | 0.442 ±0.006 | **0.380** ±0.001 | 0.436 ±0.001 | 0.412 ±0.003 |
| | 96 | **0.385** ±0.003 | 0.400 ±0.004 | 0.416 ±0.001 | **0.397** ±0.003 | **0.402** ±0.003 | 0.408 ±0.005 |
| Electricity | 24 | **0.165** ±0.001 | 0.187 ±0.003 | 0.242 ±0.007 | **0.193** ±0.001 | 0.222 ±0.001 | 0.266 ±0.001 |
| | 48 | **0.188** ±0.003 | 0.203 ±0.008 | 0.232 ±0.005 | **0.222** ±0.002 | 0.262 ±0.002 | 0.293 ±0.002 |
| | 72 | **0.209** ±0.004 | 0.230 ±0.001 | 0.263 ±0.004 | **0.238** ±0.001 | 0.280 ±0.003 | 0.310 ±0.002 |
| | 96 | **0.211** ±0.001 | 0.230 ±0.014 | 0.224 ±0.004 | **0.242** ±0.001 | 0.289 ±0.002 | 0.327 ±0.003 |
| Solar | 24 | **0.446** ±0.002 | 0.472 ±0.003 | 0.524 ±0.001 | **0.455** ±0.009 | 0.524 ±0.003 | 0.498 ±0.003 |
| | 48 | **0.546** ±0.005 | 0.603 ±0.003 | 0.622 ±0.006 | **0.556** ±0.005 | 0.629 ±0.003 | 0.690 ±0.031 |
| | 72 | **0.666** ±0.003 | **0.667** ±0.004 | 0.701 ±0.004 | **0.643** ±0.003 | 0.729 ±0.023 | 0.716 ±0.024 |
| | 96 | **0.713** ±0.004 | 0.739 ±0.009 | 0.744 ±0.002 | **0.708** ±0.004 | 0.770 ±0.004 | 0.738 ±0.002 |

**Table 10**
Comparison of our forecast-blur-denoise with standalone forecasting models with higher number of layers (parameters) denoted by † sign. Initially, the number of layers for our proposed model and other baselines are chosen from $\{1, 2\}$, however to show that the performance of our model indeed stems from its mechanism, we included the results of standalone forecasting models with number of layers chosen from $\{3, 4\}$. Results are reported as average and standard error of **MAE**. A lower **MAE** indicates a better forecasting model.

| Dataset | Horizon | AutoDG(Ours) | Autoformer | Autoformer† | InfoDG(Ours) | Informer | Informer† |
|---|---|---|---|---|---|---|---|
| Traffic | 24 | **0.333** ±0.010 | **0.334** ±0.007 | **0.332** ±0.002 | 0.355 ±0.007 | **0.329** ±0.006 | 0.382 ±0.003 |
| | 48 | **0.328** ±0.001 | 0.368 ±0.002 | 0.336 ±0.001 | **0.345** ±0.001 | 0.377 ±0.013 | 0.402 ±0.005 |
| | 72 | **0.358** ±0.013 | **0.356** ±0.003 | 0.357 ±0.001 | **0.345** ±0.003 | 0.377 ±0.010 | 0.382 ±0.011 |
| | 96 | **0.385** ±0.003 | 0.400 ±0.004 | 0.370 ±0.001 | **0.397** ±0.003 | **0.402** ±0.002 | 0.413 ±0.003 |
| Electricity | 24 | **0.249** ±0.001 | 0.265 ±0.003 | 0.303 ±0.003 | **0.290** ±0.003 | **0.300** ±0.006 | 0.332 ±0.002 |
| | 48 | **0.275** ±0.001 | 0.292 ±0.007 | 0.317 ±0.002 | **0.311** ±0.002 | 0.349 ±0.009 | 0.377 ±0.002 |
| | 72 | **0.303** ±0.004 | **0.297** ±0.007 | 0.351 ±0.002 | **0.336** ±0.003 | 0.371 ±0.002 | 0.384 ±0.002 |
| | 96 | **0.304** ±0.001 | 0.372 ±0.010 | 0.317 ±0.001 | **0.342** ±0.004 | 0.378±0.001 | 0.411 ±0.002 |
| Solar | 24 | **0.548** ±0.009 | 0.603 ±0.002 | 0.608 ±0.001 | **0.533** ±0.005 | 0.597 ±0.002 | 0.575 ±0.000 |
| | 48 | **0.612** ±0.003 | 0.656 ±0.003 | 0.672 ±0.004 | **0.624** ±0.007 | 0.681 ±0.013 | 0.745 ±0.020 |
| | 72 | **0.702** ±0.001 | 0.729 ±0.017 | 0.707 ±0.001 | **0.690** ±0.013 | 0.752 ±0.017 | 0.762 ±0.016 |
| | 96 | **0.725** ±0.000 | 0.754 ±0.009 | 0.737 ±0.002 | **0.727** ±0.006 | 0.772 ±0.012 | 0.785 ±0.010 |